# Forecasting of Indian Rupee (INR) / US Dollar (USD) Currency Exchange Rate Using Artificial Neural Network

Yusuf Perwej[1] and Asif Perwej[2]

[1]M.Tech, MCA, Department of Computer Science & Information System,
Jazan University, Jazan, Kingdom of Saudi Arabia (KSA)

`Yusufperwej@gmail.com`

[2]P.hD, MBA, Assistant Professor, Skyline Institute of Engineering & Technology
Greater Noida , U.P, India

`asifperwej@gmail.com`

## ABSTRACT

*A large part of the workforce, and growing every day, is originally from India. India one of the second largest populations in the world, they have a lot to offer in terms of jobs. The sheer number of IT workers makes them a formidable travelling force as well, easily picking up employment in English speaking countries. The beginning of the economic crises since 2008 September, many Indians have return homeland, and this has had a substantial impression on the Indian Rupee (INR) as liken to the US Dollar (USD). We are using numerational knowledge based techniques for forecasting has been proved highly successful in present time. The purpose of this paper is to examine the effects of several important neural network factors on model fitting and forecasting the behaviours. In this paper, Artificial Neural Network has successfully been used for exchange rate forecasting. This paper examines the effects of the number of inputs and hidden nodes and the size of the training sample on the in-sample and out-of-sample performance. The Indian Rupee (INR) / US Dollar (USD) is used for detailed examinations. The number of input nodes has a greater impact on performance than the number of hidden nodes, while a large number of observations do reduce forecast errors.*

## 1. INTRODUCTION

Many Indians travel abroad for work. The IT workforce from Indian is easily employed in countries like the United States, United Kingdom, Singapore and Canada. The global economic crisis that started in since 2008 September has stemmed part of this manpower flow, the circumstance is now picking up globally and Indians overseas are taking a breather, knowing that their jobs are secure and they can still send back money. Indian Rupee (INR) is partially to the US Dollar (USD), one of the key employers of Indian manpower. Currently, 1 US Dollar

DOI : 10.5121/ijcsea.2012.2204    41

(USD) is trading at 45.6 Indian Rupees. There has been so wild fluctuations since early 2009, with the exchange rate peaking at 1 USD = 52.1 INR in March 2009, to a low of 1 USD = 46.9 INR in May 2009. This is a difference of 5.2INR/US Dollar and those who remitted money from the United States back to India in March 2009 must be laughing all the way to the bank. The currency exchange rates play an important role in compulsive the dynamics of the currency market [1]. The appropriate prediction of currency exchange rate is an essential factor for the success of many businesses and investment firm. Although the market is well known for its unpredictability and volatility, there exist a number of groups (like Banks, Agency and other) for predicting exchange rates using numerous techniques. The many types of theoretical models including both econometric and time series approaches have been widely used to model and forecast exchange rates such as [2] autoregressive conditional heteroskedasticity (ARCH), general autoregressive conditional heteroskedasticity (GARCH), [3] chaotic dynamics, and self-exciting threshold autoregressive models applied to financial forecasting. While these models may be best for a particular situation they perform poorly in other applications. The Artificial Neural Networks have received increasing attention as decision-making tools.

The Artificial Neural Networks, the well-known function approximates in prediction and system modeling, has recently shown its great applicability in time series analysis and forecasting. Artificial Neural Networks assist multivariate analysis. Multivariate models can rely on grater information, where not only the lagged time series being forecast, but also other indicators (such as technical, fundamental, inter-marker etc. To financial market), are combined to act as predictors. In addition, Artificial Neural Network is more effective in describing the dynamics of non-stationary time series due to its unique non-parametric, non-assumable, noise-tolerant and adaptive properties. Artificial Neural Networks are universal function approximates that can map any nonlinear function without a priori assumptions about the data [4]. Artificial Neural Network model for forecasting exchange rates have been investigated in a number of studies have found that neural networks are better than random walk models in predicting the Deutsche mark/US dollar (DEM/USD) exchange rate [5]. In comparison with the traditional forecasting methods such as Box-Jenkins ARIMA models or regression models, [6] there are many more modeling factors to be considered in neural networks [7]. The neural network modeling issues for forecasting. Kaastra and Boyd [8] propose an eight step method in designing a neural network model for forecasting financial time series. The purpose of this research is to provide an in-depth study of the effects of several important factors on the performance of neural networks in exchange rate forecasting. Specifically we will examine two neural net- work factors .The number of input nodes and the number of hidden nodes on the forecasting performance of the exchange rate between the Indian Rupee (INR) / US Dollar (USD).

## 2. ARTIFICIAL NEURAL NETWORK

Artificial Neural Network (ANN) mimics biological information processing mechanisms. They are typically designed to perform a nonlinear mapping from a set of inputs to a set of outputs. Artificial Neural Network is developed to try to achieve the biological system type of performance using a dense interconnection of simple processing [9] elements analogous to biological neurons. The An artificial neural network is a biologically inspired computational model which consists of processing elements (called neurons) and connections among them with coefficients (weights) bound to the connections, which constitute the neuronal structure, and training and recall algorithm attached to the structure. Neural Network is called connectionist models because of the main role of the connections on them. The connection weights are the "memory" of the system. Even though neural Network has similarities to the human brain, they are not meant to model it. They are meant to be useful models for problem-solving and knowledge-engineering in a "humanlike" way. The human brain is much more complex and unfortunately, many of its cognitive functions are still not well known. But the more we learn about the human brain, the better computational models are developed and put to practicable use. The human brain contains about $10^{11}$ neurons participating in perhaps $10^{15}$ interconnections over transmission paths. Each of these paths could be a meter long or more.

## 2. 1 Neurons

Neurons share many characteristics with the other cells in the body, but they have unique capabilities for receiving, processing, and transmitting electrochemical signals over the neural pathways that make up the brain's communication system [10]. Each input has its own relative weight which gives the input the impact that it needs on the processing element's summation function. These weights perform the same type of function as do the varying synaptic strengths of biological neurons. In both cases, some inputs are made more important than others so that they have a greater effect on the processing element as they combine to produce a neural response. Weights are adaptive coefficients within the network that determine the intensity of the input signal as registered by the artificial neuron.

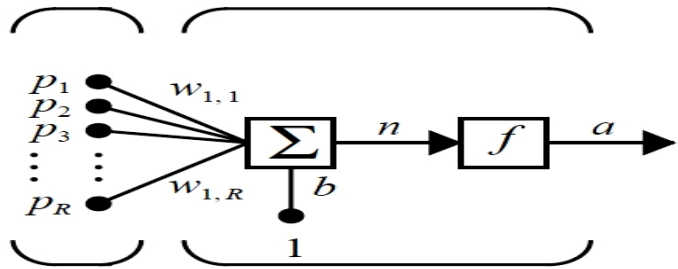

Figure - 1 A simple neuron with R inputs

They are a measure of an input's connection strength. These strengths can be modified in response to various training sets and according to a network specific topology or through its learning rules. The vast majority of artificial neural network solutions have been trained in supervision. In this mode, the actual output of a neural network is compared to the desired output. Weights, which are usually randomly set to begin with, are then adjusted by the network so that the next iteration, or cycle, will produce a closer match between the desired and the actual output. The learning method tries to minimize the current errors of all processing elements. This global error reduction is created over time by continuously modifying the input weights until acceptable network accuracy is reached.

| Hard limit | Pure linear | Sigmoid | Tansigmoid |
|---|---|---|---|
| $f(x) = \begin{cases} 1, x \geq 0 \\ 0, x < 0 \end{cases}$ | $f(x) = x$ | $f(x) = \dfrac{1}{1+e^{-x}}$ | $f(x) = \dfrac{2}{1+e^{-2n}} - 1$ |
| $f(x) \in \{0,1\}$ | $f(x) \in (-\infty, +\infty)$ | $f(x) \in [0,1]$ | $f(x) \in [-1,1]$ |

Table - 1 The most commonly used Transfer functions

## 2. 2 Learning

Learning in artificial neural network is achieved by varying the connection weights iteratively so that the network is trained to perform certain tasks. It generally involves the minimization of some error function say the total mean square error between the actual and expected output under the supervision of a trainer. This is often called supervised training. However, in some cases, the exact desired output is not known. Reinforcement learning is used in such cases and training is based only on whether the actual output is correct or not. Unsupervised learning tries to find correlations among input data when no information on the correctness of the output is available [11]. The rule followed to update the connection weights the learning rule determines how well the network converges towards its desired optimality.

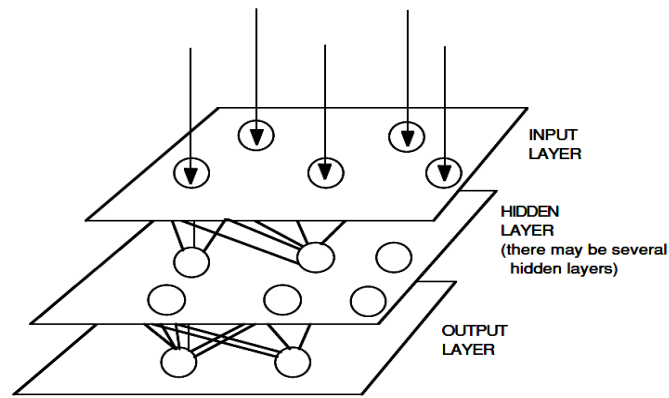

Figure - 2 Neural Network Diagram

## 2. 3 MLP

A multilayer perceptrons are a network of simple neurons called perceptrons. The basic concept of a single perceptron was introduced by Rosenblatt in 1958. The perceptron computes a single output from multiple real-valued inputs by forming a linear combination according to its input weights and then possibly putting the output through some nonlinear activation function. A single perceptron is not very useful because of its limited mapping ability. No matter what activation function is used, the perceptron is only able to represent an oriented ridge-like function. The perceptrons can, however, be used as building blocks of a larger, much more practicable structure. A typical multilayer perceptron (MLP) network consists of a set of source nodes forming the input layer, one or more hidden layers of computation nodes, and an output layer of nodes. The input signal propagates through the network layer-by-layer.

## 3. ARTIFICIAL NEURAL NETWORK FOR TIME SERIES FORECASTING

Time series forecasting is highly utilized in predicting economic and business trends .Recently, artificial neural networks that serve, as a powerful computational framework, have gained much popularity in business applications. Artificial Neural Networks have been successfully applied to loan evaluation, signature recognition, time series forecasting, classification analysis and many other difficult pattern recognition problems [12]. Artificial Neural Networks models have been proposed and used for the forecasting purpose. The most popular and successful one is the feed forward multilayer network or the multilayer perceptron (MLP). The MLP is typically a combination of several layers of nodes. The first lowest layer is input layer where external information is received. The last or the highest layer is an output layer where the problem solution is obtained. The input layer and output layer are separated by one or more intermediate layers called the hidden layers. The nodes in adjacent layers are usually fully connected by acyclic arcs from a lower layer to a higher layer [13].

The knowledge learned by a network is stored in the arcs and the nodes in the form of arc weights and node biases which will be estimated in the neural network training process. Figure 3 is an example of a fully connected MLP with one hidden layer. For a univariate time series forecasting problem, the inputs of the network are the past, lagged observations [14] of the data series and the outputs are the future values. Each input pattern is composed of a moving window of fixed length along the series. The network representation is a mapping function of the form

$$y_{t+1} = f(y_t, y_{t-1}, \dots, y_{t-p})$$

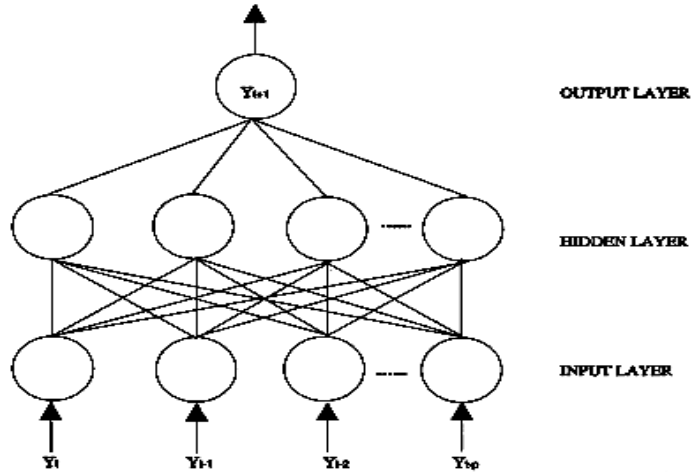

Figure - 3 A typical fully connected feed forward neural network for time series forecasting

We are notice that the here $y_t$ is the contemplation at time t as will as p is the dimension of the input value in other words the number of past contemplation related to the future value. The first training pattern is composed of $y_1, y_2, \dots, y_p$ as the inputs and $y_{p+1}$ as the target output. The second training pattern contains $y_1, y_2, \dots, y_{p+1}$ for the inputs and $y_{p+2}$ for the desired output. Finally, the last training pattern is $y_{N-p}, y_{N-p+1}, \dots, y_{N-1}$ for the inputs and $y_N$ for the target. The neural network training objective is to find the weights in order that some overall predictive error measure such as the sum of the squared errors (SSE) is minimized [15]. In this network structure, SSE can be written as

$$SSE = \sum_{i=p+1}^{N} (y_i - a_i)^2$$

Where $a_i$ is the output from the network.

## 4. PROPOSED DESIGEN

The exchange rates between the Indian Rupee (INR) / US Dollar (USD) are obtained from Datastream International [16]. Data is composed of daily rates from the beginning of 1989 through the end of 2009. These papers examine the effects of several factors on the in sample fit and out of sample forecasting capabilities of neural networks. The neural network factors investigated are the number of inputs and hidden nodes which are two critical parameters in the design of a neural network. The number of input nodes is perhaps the most important factor in neural network analysis of a time series since it corresponds to the number of past lagged observations related to future values. It also plays a role in determining the autocorrelation structure of a time series. The number of hidden nodes allows neural networks to capture nonlinear patterns and detect complex relationships in the data.

We experiment with a relatively large number of input nodes. There is no upper limit on the possible number of hidden nodes in theory. However, it is rarely seen in the literature that the number of hidden nodes is more than doubling the number of input nodes .In addition, previous research [17] indicates that the forecasting performance of neural networks is not as sensitive to the number of hidden nodes as it is to the number of input nodes. Thus, five levels of hidden nodes, 6, 12, 18, 24 and 30 will be experimented. The combination of ten input nodes and five hidden nodes yields a total of 50 neural network architectures being considered for each in-sample training data set. A comprehensive study of neural network time series forecasting, finds that neural network models do not necessarily require large data [18] set to perform well. To test if there is a significant difference between large and small training samples in modelling and forecasting exchange rates, we use two training sample sizes in our study. The large sample consists of 1043 observations from 1989 to 2009 and the small one includes 365 data points from 2003 to 2009. The test sample for both cases is the 2010 data which has 52 observations. The random walk model will be used as a benchmark for comparison. The random walk is a one-step-ahead forecasting model since it uses the current observation to predict the next one

Three-layer feed forward neural networks are used to forecast the Indian Rupee (INR) / US Dollar (USD) exchange rate. Logistic activation functions are employed in the hidden layer and the linear activation function is utilized in the output layer. We are interested in one-step-ahead forecasts, one output node are deployed in the output layer. The of using of direct optimization procedure in neural network training. To be more certain of getting the true global optima, a common practice is to solve the problem using a number of randomly generated initial solutions. We train each network 50 times by using 50 sets of different initial arc weights. The best solution among 50 runs is used as the optimal neural network training solution. The forecasting performance of the model is evaluated against a number of widely used statistical metric, we use three popular metrics RMSE, MAE, and MAPE, to evaluate the predictive performance of neural networks. These forecasting accuracy measures are listed as follows.

$$RMSE = \sqrt{\frac{\sum (y_t - \hat{y}_t)^2}{T}}$$

$$MAE = \frac{\sum |y_t - \hat{y}_t|}{T}$$

$$MAPE = \frac{1}{T}\sum \left|\frac{y_t - \hat{y}_t}{y_t}\right| \times 100$$

Where $y_t$ the actual observation is $\hat{y}_t$ is the predicted value, and T is the number of predictions. These criteria are mean based and are frequently used performance measures in this paper.

## 5. EXPERIMENTAL RESULTS OF FORECASTING THE INDIAN RUPEE (INR) / US DOLLER (USD)

The goal of this paper has been to study is to investigate the effects of neural network factors on the modelling and forecasting performance of neural networks, both in sample (training set) and out-of-sample (test set) results. The focus will be on the out-of-sample analysis because it is the forecasting capability that researchers and practitioners are most interested in. For neural network method, due to the potential problem of over fitting, we need to study the conditions under which over fitting may occur. An over fitted model gives good in-sample fit to the training data, yet poor predictive out-of-sample performance. Therefore, the examination of both in- sample and out-of-sample behaviours will provide us information on when and how over fitting occurs.

| Input | Hidden | RMSE | MAE | MAPE |
|---|---|---|---|---|
| 1 | 6 | 0.08493210 | 0.08452030 | 2.85874590 |
| 1 | 12 | 0.08493210 | 0.08451020 | 2.85871860 |
| 1 | 18 | 0.08493210 | 0.08451020 | 2.85872360 |
| 1 | 24 | 0.08493200 | 0.08451020 | 2.85873890 |
| 1 | 30 | 0.08493200 | 0.08451020 | 2.85872910 |
| Avgr | | 0.08493210 | 0.08451020 | 2.85873120 |
| 2 | 6 | 0.08493190 | 0.08449910 | 2.85860740 |
| 2 | 12 | 0.08493120 | 0.08449950 | 2.85951250 |
| 2 | 18 | 0.08493110 | 0.08449940 | 2.85913650 |
| 2 | 24 | 0.08493080 | 0.08449970 | 2.85926800 |
| 2 | 30 | 0.08493080 | 0.08449950 | 2.85912570 |
| Avgr | | 0.08493120 | 0.08449940 | 2.85913000 |
| 3 | 6 | 0.08483930 | 0.08449790 | 2.85828000 |
| 3 | 12 | 0.08483910 | 0.08449840 | 2.85966620 |
| 3 | 18 | 0.08483820 | 0.08449890 | 2.85995800 |
| 3 | 24 | 0.08483810 | 0.08449730 | 2.85881060 |
| 3 | 30 | 0.08483800 | 0.08449830 | 2.85912780 |
| Avgr | | 0.08483850 | 0.08449820 | 2.85916850 |
| 4 | 6 | 0.08490000 | 0.08449680 | 2.85934930 |
| 4 | 12 | 0.08483780 | 0.08449550 | 2.85750980 |
| 4 | 18 | 0.08483650 | 0.08449650 | 2.85932080 |
| 4 | 24 | 0.08483570 | 0.08449550 | 2.85755380 |
| 4 | 30 | 0.08483440 | 0.08449710 | 2.85933500 |
| Avgr | | 0.08483690 | 0.08449630 | 2.85861370 |
| 5 | 6 | 0.08483720 | 0.08449780 | 2.85864780 |
| 5 | 12 | 0.08483500 | 0.08449550 | 2.85682010 |
| 5 | 18 | 0.08483360 | 0.08449500 | 2.85720080 |
| 5 | 24 | 0.08483030 | 0.08449640 | 2.85868170 |
| 5 | 30 | 0.08473780 | 0.08449240 | 2.85635120 |
| Avgr | | 0.08483280 | 0.08449540 | 2.85754030 |
| 6 | 6 | 0.08483510 | 0.08449920 | 2.85790900 |
| 6 | 12 | 0.08483210 | 0.08449930 | 2.85762540 |
| 6 | 18 | 0.08483900 | 0.08449230 | 2.85476370 |
| 6 | 24 | 0.08473730 | 0.08449750 | 2.85814660 |
| 6 | 30 | 0.08473370 | 0.08449520 | 2.85655250 |
| Avgr | | 0.08473950 | 0.08449670 | 2.85699940 |
| 7 | 6 | 0.08483600 | 0.08449990 | 2.86019840 |
| 7 | 12 | 0.08483190 | 0.08450040 | 2.86007590 |
| 7 | 18 | 0.08483120 | 0.08449590 | 2.85815280 |
| 7 | 24 | 0.08473700 | 0.08449770 | 2.85879190 |
| 7 | 30 | 0.08473570 | 0.08449650 | 2.85752230 |
| Avgr | | 0.08483030 | 0.08449810 | 2.85894830 |
| 8 | 6 | 0.08483480 | 0.08450020 | 2.85772100 |
| 8 | 12 | 0.08483250 | 0.08449870 | 2.85956610 |
| 8 | 18 | 0.08473980 | 0.08449910 | 2.85873100 |
| 8 | 24 | 0.08473730 | 0.08449600 | 2.85672850 |
| 8 | 30 | 0.08473480 | 0.08449650 | 2.85537220 |
| Avgr | | 0.08473990 | 0.08449810 | 2.85762380 |
| 9 | 6 | 0.08483430 | 0.08449990 | 2.85891170 |
| 9 | 12 | 0.08483350 | 0.08449930 | 2.85875390 |
| 9 | 18 | 0.08473540 | 0.08449830 | 2.85776730 |
| 9 | 24 | 0.08473520 | 0.08449590 | 2.85696390 |
| 9 | 30 | 0.08463750 | 0.08449000 | 2.85134280 |
| Avgr | | 0.08473720 | 0.08449670 | 2.85674790 |
| 10 | 6 | 0.08473700 | 0.08449230 | 2.85370770 |
| 10 | 12 | 0.08473730 | 0.08449400 | 2.85579050 |
| 10 | 18 | 0.08463990 | 0.08448860 | 2.85245050 |
| 10 | 24 | 0.08473150 | 0.08449480 | 2.85637290 |
| 10 | 30 | 0.08463900 | 0.08449450 | 2.85531560 |
| Avgr | | 0.08473290 | 0.08449280 | 2.85472740 |

Table - 2 Artificial Neural Network factors on training performance (training period 1989 - 2009) of effects

The table 2 shows the in-sample results for the large training sample of 1043 observations. It is quite evident that as the number of hidden nodes increases, RMSE decreases. This pattern is observed consistently in each level of the input node. The more hidden nodes are used the neural network becomes more powerful in modelling the data. We observe a different pattern for the effects of the input node and the hidden node with MAE and MAPE. MAE and MAPE do not decrease in general as the number of hidden nodes increases within each level of the input node. The number of input nodes increases from 1 to 5, the mean MAE steadily decreases from 0.08451020 to 0.08449540 .When the number of input nodes is in the range of 6 to 10, overall MAE increases first and then decreases. MAPE does not show a clear input node effect as will as MAE and MAPE is doesn't minimize but RMSE is minimize using proposed artificial neural networks model. It is important to note that there is less variation between different hidden node levels within each input node level than among different input node levels, suggesting that the number of input nodes has greater impact on the model fitting process of Artificial Neural Networks. The out-of- sample analysis to examine the predictive capabilities of Artificial Neural Networks as the Artificial Neural Network structure changes Artificial Neural Network performance will also be compared with that of the forecasting models selected by SCA which happen to be the random walk model in our paper.

| Input | RMSE1 | MAE1 | MAPE1 |
|---|---|---|---|
| 1 | 0.08460500 | 0.08499800 | 2.87116440 |
| 2 | 0.08460770 | 0.08505200 | 2.87409740 |
| 3 | 0.08458220 | 0.08487500 | 2.86452700 |
| 4 | 0.08457100 | 0.08477800 | 2.85934250 |
| 5 | 0.08455450 | 0.08437900 | 2.83804880 |
| 6 | 0.08449670 | 0.08382900 | 2.80883310 |
| 7 | 0.08452120 | 0.08414400 | 2.82503450 |
| 8 | 0.08451160 | 0.08406800 | 2.82107130 |
| 9 | 0.08454350 | 0.08420200 | 2.82859520 |
| 10 | 0.08455370 | 0.08429200 | 2.83344960 |
| RW | 0.08460700 | 0.08501600 | 2.87208070 |

| Input | RMSE2 | MAE2 | MAPE2 |
|---|---|---|---|
| 1 | 0.08475000 | 0.08601800 | 3.84724010 |
| 2 | 0.08475250 | 0.08602600 | 3.84758330 |
| 3 | 0.08477910 | 0.08598300 | 3.84535810 |
| 4 | 0.08475080 | 0.08596100 | 3.84403970 |
| 5 | 0.08475330 | 0.08594100 | 3.84350680 |
| 6 | 0 08475530 | 0.08584300 | 3.83823520 |
| 7 | 0.08476640 | 0.08597800 | 3.84555150 |
| 8 | 0.08477240 | 0.08600100 | 3.84643580 |
| 9 | 0.08478300 | 0.08604700 | 3.84936020 |
| 10 | 0.08478270 | 0.08615600 | 3.85593760 |
| RW | 0.08475220 | 0.08604500 | 3.84863850 |

| Input | RMSE3 | MAE3 | MAPE3 |
|---|---|---|---|
| 1 | 0.08467120 | 0.08533400 | 3.80582120 |
| 2 | 0.08467220 | 0.08533900 | 3.80606810 |
| 3 | 0.08467020 | 0.08531500 | 3.80479330 |
| 4 | 0.08467100 | 0.08530100 | 3.80396130 |
| 5 | 0.08467170 | 0.08530400 | 3.80446360 |
| 6 | 0.08467440 | 0.08528500 | 3.80367020 |
| 7 | 0.08467980 | 0.08533700 | 3.80648640 |
| 8 | 0.08468160 | 0.08530600 | 3.80457480 |
| 9 | 0.08468890 | 0.08536100 | 3.80787970 |
| 10 | 0.08469180 | 0.08542800 | 3.81184880 |
| RW | 0.08467300 | 0.08536000 | 3.80720790 |

Table - 3 Artificial Neural Network effects of input nodes (training period 1989 - 2009) Out-of-sample analysis

The prediction of out-of-sample results from using the large training sample contains a table 3. The using 1, 2 and 3 for each performance measure of the three time horizons. That is RMSE1, MAE1 and MAPE1 are used for the one-month time horizon and RMSE2, MAE2 and MAPE2 are for the six-month time horizon and RMSE3, MAE3, and MAPE3 are for the 12-month horizon. The one-month horizon, all three measures of performance indicate 6 input nodes produce the best predictions. Average RMSE, MAE and MAPE take on values of 0.08449670, 0.08382900 and 2.80883310 respectively. As the length of time horizon increases, the effects of input nodes on MAE and MAPE are quite consistent. The network with six input nodes is still the overall best architecture. The observed pattern in RMSE for 6- and 12-month horizons is not the same as in the case of one-month horizon. For 6- and 12-month time horizons, the lowest average RMSE is 0.08477910 and 0.08467020 which occur with three input nodes, indicating that the specification of the number of input nodes may be sensitive to the performance measure and the forecast horizon. The random walk model in three performance measures across the three time horizons are reported at the bottom of table 3. It is clear that neural networks predict much better than the random walk model in terms of all three measures across the three time horizons not only for the best neural network models but also for most other network architectures.

| Sample Size | Input | Hidden | RMSE1 | MAE1 | MAPE1 |
|---|---|---|---|---|---|
| Large | 5 | 6 | 0.08453490 | 0.08410600 | 2.82343980 |
| | 5 | 12 | 0.08448200 | 0.08363700 | 1.89774230 |
| | 5 | 18 | 0.08452950 | 0.08421500 | 2.82911360 |
| | 5 | 24 | 0.08445390 | 0.08342300 | 1.88651750 |
| | 5 | 30 | 0.08449820 | 0.08380700 | 2.80734910 |
| Small | 1 | 6 | 0.08459730 | 0.08456300 | 2.84932490 |
| | 1 | 12 | 0.08459730 | 0.08456100 | 2.84927040 |
| | 1 | 18 | 0.08459740 | 0.08456200 | 2.84931450 |
| | 1 | 24 | 0.08459740 | 0.08456400 | 2.84939970 |
| | 1 | 30 | 0.08459730 | 0.08456100 | 2.84925390 |

| Sample Size | Input | Hidden | RMSE2 | MAE2 | MAPE2 |
|---|---|---|---|---|---|
| Large | 5 | 6 | 0.08475230 | 0.08595400 | 3.84428700 |
| | 5 | 12 | 0.08475680 | 0.08584700 | 3.83832860 |
| | 5 | 18 | 0.08475030 | 0.08587800 | 3.83963800 |
| | 5 | 24 | 0.08476450 | 0.08577100 | 3.83463120 |
| | 5 | 30 | 0.08475240 | 0.08576500 | 3.83429130 |
| Small | 1 | 6 | 0.08496140 | 0.08749300 | 4.84484470 |
| | 1 | 12 | 0.08496670 | 0.08754000 | 4.84778130 |
| | 1 | 18 | 0.08496510 | 0.08752600 | 4.84688520 |
| | 1 | 24 | 0.08496270 | 0.08750500 | 4.84558060 |
| | 1 | 30 | 0.08496590 | 0.08753300 | 4.84737830 |

| Sample Size | Input | Hidden | RMSE3 | MAE3 | MAPE3 |
|---|---|---|---|---|---|
| Large | 5 | 6 | 0.08467580 | 0.08535600 | 3.80763220 |
| | 5 | 12 | 0.08467540 | 0.08528800 | 3.80371900 |
| | 5 | 18 | 0.08467260 | 0.08530200 | 3.80426750 |
| | 5 | 24 | 0.08468000 | 0.08529300 | 3.80440940 |
| | 5 | 30 | 0.08466840 | 0.08518500 | 2.89832270 |
| Small | 1 | 6 | 0.08486520 | 0.08693700 | 4.80652020 |
| | 1 | 12 | 0.08486960 | 0.08698100 | 4.80918540 |
| | 1 | 18 | 0.08486830 | 0.08696800 | 4.80840140 |
| | 1 | 24 | 0.08486640 | 0.08695000 | 4.80727580 |
| | 1 | 30 | 0.08486890 | 0.08697500 | 4.80878610 |

Table - 4 Artificial Neural Network Effects of hidden nodes

The effects of hidden nodes on the out-of-sample performance for both the large and small samples contain table 4. Because of the similarity in the effects with different levels of input nodes, only the results with five input nodes for the large sample and one input node for the small sample are reported. Results show no clear effects of hidden nodes across different time horizons along different performance measures. The differences in performance measures across the levels of hidden nodes are very small, indicating the number of hidden nodes does not occupy an important role in the out-of-sample performance of neural networks. We observe that within each input node level, the results are not very sensitive to the change in the number of hidden nodes. Hence, correctly identifying the number of input nodes is more important than identifying the number of hidden nodes. Our results are in line with the findings reported by Lachtermacher and Fuller [15] who apply neural networks to predict one-step-ahead river flow The performance measures for the neural network model in the 1989 2009 series (the large training sample) take on larger values. This suggests that the Artificial Neural Networks is a better choice for long term forecasting.

## 6. CONCLUSIONS

Finally, the forecasting of Exchange rates using an Artificial Neural Network model .In this paper we investigate the effects of three important factors on Artificial Neural Networks modelling and forecasting performance for Indian Rupee (INR) / US Dollar (USD) exchange rate. The number of input nodes and the number of hidden nodes are the experimental Artificial Neural Network factors. Both in-sample fitting ability and out- of-sample predictive performance with three forecast horizons are evaluated along three criteria, RMSE, MAE, and MAPE. The effects of two training sample sizes are examined with the identical forecast horizons. The purpose of this paper is to examine the effects of several important neural network factors on model fitting and forecasting the behaviours. The forecasting purpose it is not appropriate to evaluate the Artificial Neural Network capability with the training sample alone. There are no broadly accepted methods for construction of the best predictive model using strictly in-sample training data. The selection of the optimal network architecture should be based on test sample results. We have also found that the number of input nodes plays an important role in neural network time series forecasting. In this paper we are studying in the Indian Rupee (INR) / US Dollar (USD) exchange rate is selected for circumstantial analysis. The many other studies in exchange rate forecasting show that there is little difference in in-sample fitting and out-of-sample predictive results from one exchange rate to another. We investigate only one-step-ahead forecasting strategy. Robustness of neural networks to the changing structures, it can easily handle the inaccuracy and any degree of nonlinearity in the data.